\documentclass[10pt,twocolumn,letterpaper]{article}

\usepackage{iccv}
\usepackage{times}
\usepackage{epsfig}
\usepackage{graphicx}
\usepackage{subfigure}
\usepackage{amsmath}
\usepackage{amssymb}
\usepackage{float}
\usepackage{algorithm}
\usepackage{algorithmic}
\usepackage{multirow}
\usepackage{booktabs,tabularx}
\usepackage{enumitem}
\usepackage{makecell}
\usepackage{indentfirst}
\usepackage{authblk}

\usepackage[pagebackref=true,breaklinks=true,letterpaper=true,colorlinks,bookmarks=false]{hyperref}

\iccvfinalcopy 

\def\iccvPaperID{****} 

\ificcvfinal\fi

\newcommand{\fig}[1]{Figure \ref{#1}}

\newcommand{\tab}[1]{Table \ref{#1}}

\newcommand*\samethanks[1][\value{footnote}]{\footnotemark[#1]}

\begin{document}

\title{Decoupling Visual-Semantic Feature Learning for Robust \\Scene Text Recognition}

\author[12]{Changxu Cheng \thanks{Equal contribution.}}
\author[1]{Bohan Li \samethanks[1]}
\author[2]{Qi Zheng}
\author[2]{Yongpan Wang}
\author[1]{Wenyu Liu}
\affil[1]{Huazhong University of Science and Technology, China}
\affil[2]{Alibaba Group, China}
\affil[ ]{\tt\small \{cxcheng,bohan1024,liuwy\}@hust.edu.cn, \tt\small \{yongqi.zq,yongpan\}@taobao.com}

\maketitle
\ificcvfinal\thispagestyle{empty}\fi

\begin{abstract}
   Semantic information has been proved effective in scene text recognition. Most existing methods tend to couple both visual and semantic information in an attention-based decoder. As a result, the learning of semantic features is prone to have a bias on the limited vocabulary of the training set, which is called vocabulary reliance. In this paper, we propose a novel Visual-Semantic Decoupling Network (VSDN) to address the problem. Our VSDN contains a Visual Decoder (VD) and a Semantic Decoder (SD) to learn purer visual and semantic feature representation respectively. Besides, a Semantic Encoder (SE) is designed to match SD, which can be pre-trained together by additional inexpensive large vocabulary via a simple word correction task. Thus the semantic feature is more unbiased and precise to guide the visual feature alignment and enrich the final character representation. Experiments show that our method achieves state-of-the-art or competitive results on the standard benchmarks, and outperforms the popular baseline by a large margin under circumstances where the training set has a small size of vocabulary.
\end{abstract}

\section{Introduction}

Text carries rich semantic information that is useful in many practical applications such as automatic driving, intelligent transportation system, scene understanding and so on. Reading text in scene images plays an important role in artificial intelligence. 

\begin{figure}
   \centering
   \includegraphics[scale=0.4]{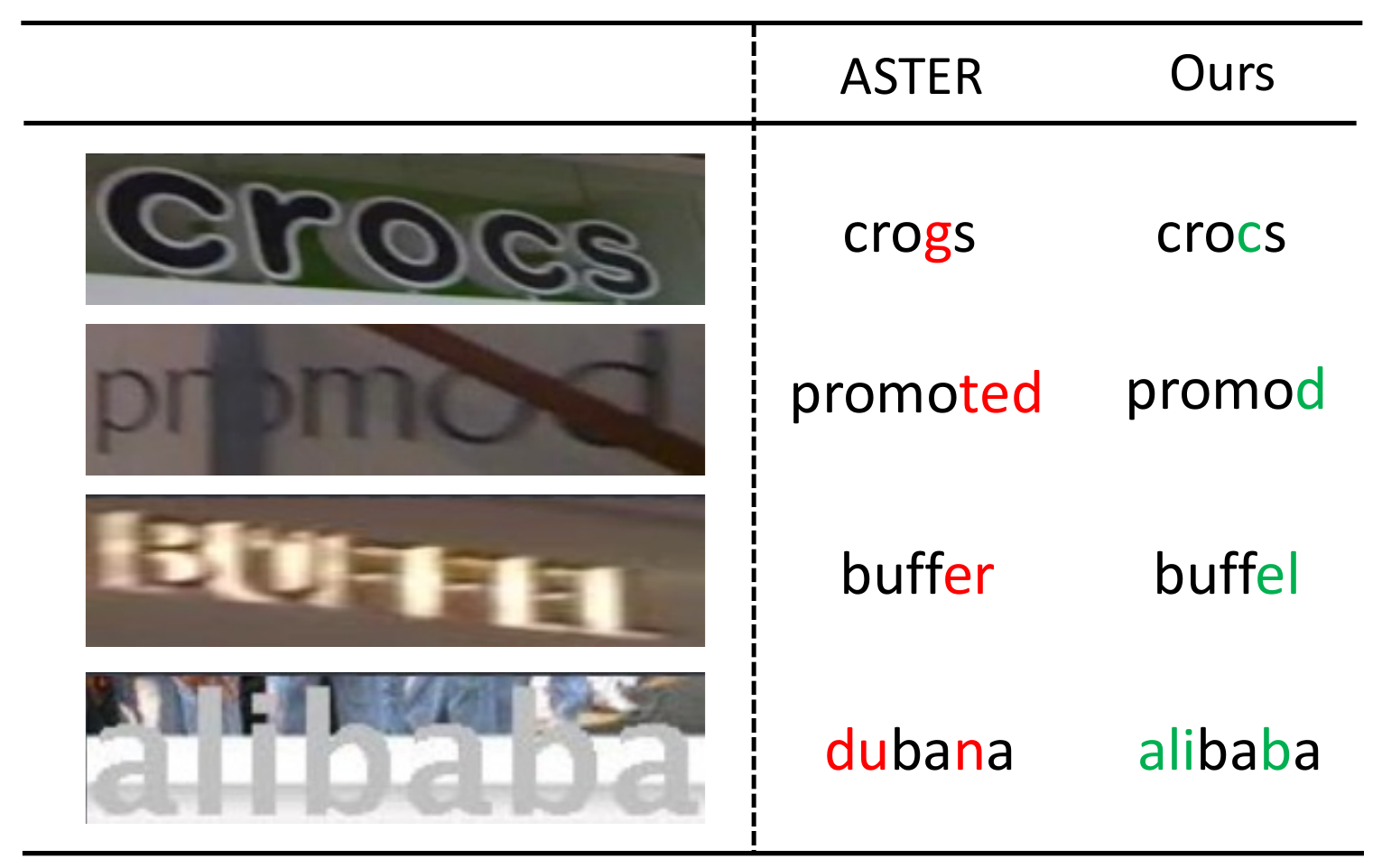}
   \caption{The comparison of VSDN and the V-S coupling method like Aster~\cite{DBLP:journals/pami/ShiYWLYB19}. Ours performs better on the cases when the words are out of training vocabulary.}
   \label{cmp_cases}
\end{figure}

\begin{figure}
   \centering
   \includegraphics[scale=0.45]{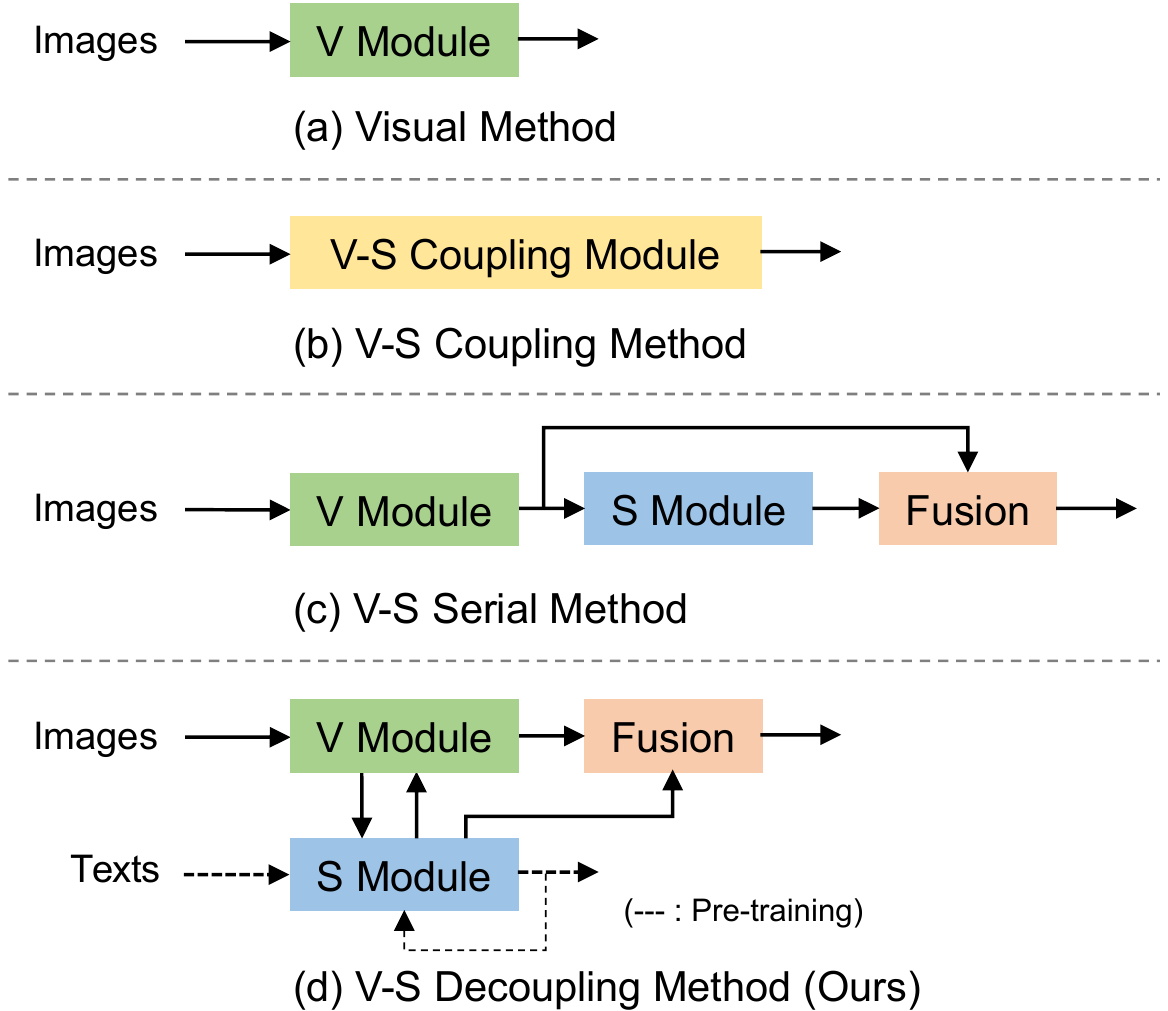}
   \caption{Different kinds of scene text recognition methods from the perspective of visual and semantic learning. 'V' and 'S' denote 'visual' and 'semantic' respectively.}
   \label{cmp_methods}
\end{figure}

In the community of scene text recognition, semantic information has been shown useful along with the visual feature in an end-to-end model~\cite{DBLP:journals/ijon/ChenWZJL20,DBLP:conf/cvpr/QiaoZYZ020,DBLP:conf/cvpr/YuLZLHLD20,DBLP:conf/mm/ZhengQWB20}, especially in the case of blurred or occluded text images. However, the learning of semantic feature is likely to be a double-edged sword, where the side effect is called vocabulary reliance~\cite{DBLP:conf/cvpr/WanZZLY20}. The previous visual-semantic coupling method (\fig{cmp_methods}(b)) suffers much from the effect that the model performs well on images with words within vocabulary of the training set but generalizes poorly on images with words outside vocabulary. As shown in \fig{cmp_cases}, the existing V-S coupling methods like ASTER~\cite{DBLP:journals/pami/ShiYWLYB19} incline to misrecognize texts as words that have appeared in the training phase.

The vocabulary reliance effect is mainly caused by that these V-S coupling methods learn visual and semantic feature in a hybrid decoder simultaneously. The semantics is only learned from the limited and noisy word set in the training image data, thus goes overfitting and inaccurate under a parameter-rich decoder. Hence, the processes of character alignment and character representation is poisoned by the wrongly learned semantics.


To address the problem, we propose the Visual-Semantic Decoupling Network (VSDN), in which the learning processes of character-level visual and semantic feature are realized in the visual decoder and the semantic decoder respectively. Besides, a bidirectional semantic encoder (SE) is designed to match the semantic decoder (SD). SE and SD makes up a semantic module that can be pre-trained with a word correction task to gain extra semantic information from inexpensive text data (\fig{cmp_methods}(d)). As a result, the extracted character-level semantic feature is more accurate and robust, improving the character feature alignment and representation.

Succinctly, the main contributions of this paper are three-fold. Firstly, we propose a novel Visual-Semantic Decoupling Network (VSDN) that decouples visual and semantic feature learning, which alleviates vocabulary reliance problem.
Secondly, we design a character-level semantic module, which can be easily pre-trained with a word correction task and further initialize the corresponding part in our VSDN.
Thirdly, the experiments conducted on several public datasets demonstrate that our proposed method achieves state-of-the-art or competitive recognition performance under fair comparison. Specially, the performance under circumstance of lack-of-words and lack-of-images is very surprising, compared with the visual-semantic coupled methods.

\section{Related Works}

As depicted in \fig{cmp_methods}, scene text recognition methods can be divided into three categories by the way modeling semantic information: visual method, visual-semantic coupling method and visual-semantic serial method.

\textbf{Visual Method}: Treating scene text recognition as a purely visual task, visual method is semantic-free. ~\cite{DBLP:journals/ijcv/JaderbergSVZ16} directly classifies a given text image into one of the pre-defined 90k word classes and thus is incapable of coping with those text images with words out of the pre-defined word lexicon. ~\cite{DBLP:journals/pami/ShiBY17} uses CNN and RNN to encode a given text image into a sequence feature and then feed it into a CTC~\cite{DBLP:conf/icml/GravesFGS06} decoder to align each character at each time step. Inspired by the success of visual segmentation, ~\cite{DBLP:conf/aaai/LiaoZWXLLYB19} propose a segmentation-based approach for STR, which uses FCN to predict each pixel's character class and gather the characters into a word. ~\cite{DBLP:conf/aaai/WanHCBY20} is segmentation-based as well, which represents the position and order of characters with different channels to better align characters. These segmentation-based methods need expensive character-level annotations.

\textbf{Visual-Semantic Coupling Method}: Semantic information plays a complementary role when visual information is insufficient due to the low quality of images. After encoding input text image into a 1D sequence feature, ~\cite{DBLP:conf/cvpr/LeeO16} decodes the sequence feature into target sequence with the attention mechanism~\cite{DBLP:journals/corr/BahdanauCB14}. At each time step, the result of the last time step will impact on the present result, and the semantic information that contains the dependency between characters is built during this process. ~\cite{DBLP:journals/pami/ShiYWLYB19} applies a similar approach, which adds a rectification module before CNN to alleviate the difficulty brought by the spatial layout of text images. ~\cite{DBLP:conf/iccv/YangGLHBBYB19} utilizes a symmetry-constrained rectification module which has a better performance for rectifying text images. 
~\cite{DBLP:conf/cvpr/ZhanL19} improves rectification performance by iteratively rectifying text images. ~\cite{DBLP:conf/cvpr/ChengXBNPZ18} extracts image features in four directions and fused them using a filter gate. 
Inspired by ~\cite{DBLP:conf/interspeech/HoriWZC17}, which combines CTC with attention mechanism on speech recognition tasks, ~\cite{DBLP:journals/access/ZuoSMQJ19} uses a CTC-Attention mechanism to gain better performance. Since CTC decoder has an advantage in inference speed but attention-based decoder is better at learning good feature representations, ~\cite{DBLP:conf/aaai/HuCHYL20} proposes to learn feature representations with powerful attentional guidance for better performance while using a CTC decoder to maintain a fast inference speed. ~\cite{DBLP:journals/ijon/ChenWZJL20} uses a gate to control the influence of the last time step's semantic information on the present time step. ~\cite{DBLP:conf/cvpr/QiaoZYZ020} uses the word embedding from a pre-trained
language model to predict additional global semantic information to guide the
decoding process. Though achieving good performance, these attention-based methods build visual and semantic information in a coupling way since they use one decoder to decode visual and semantic features simultaneously and thus have vocabulary reliance problem. 

\textbf{Visual-Semantic Serial Method}: ~\cite{DBLP:conf/cvpr/YuLZLHLD20} uses two modules in serial structure to decode visual features and semantic features separately. When decoding in the semantic module, ~\cite{DBLP:conf/cvpr/YuLZLHLD20} utilizes the transformer unit~\cite{DBLP:conf/nips/VaswaniSPUJGKP17} to build global semantic information. This serial way makes ~\cite{DBLP:conf/cvpr/YuLZLHLD20} largely depends on the visual module's output, especially the feature alignment, which is generated without utilizing the latter semantic information.

\section{Methodology}\label{method}

\begin{figure*}
\begin{center}
   \includegraphics[scale=0.4]{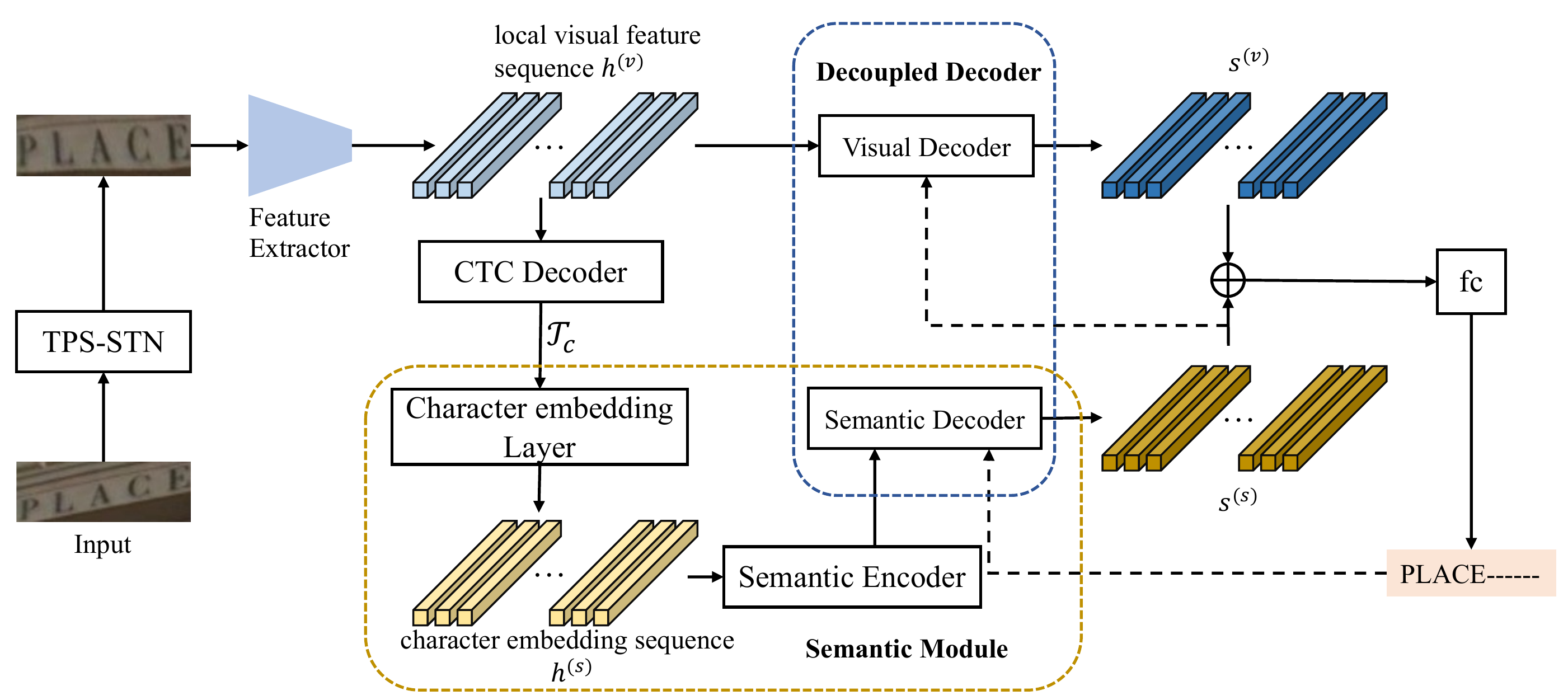}
\end{center}
\caption{Architecture of the proposed Visual-Semantic Decoupling Network. After extracting the local visual features, we exploit a visual decoder and a semantic module (semantic encoder and decoder) respectively to get character-level visual and semantic features.  The dashed line means the recurrent progress.}
\label{pipeline}
\end{figure*}

The overall structure of our proposed network is illustrated in \fig{pipeline}. It comprises four components: (1) a shared feature extractor that takes a rectified image as input and encodes it into a local visual feature sequence; (2) an attention-based visual decoder that outputs  the character-level visual feature sequence; (3) a semantic module that consists of a character embedding layer, a semantic encoder and a semantic decoder that outputs character-level semantic feature sequence; (4) a fusion block that combines the character-level visual and semantic feature representation to get the final recognition results.

Given a rectified image $I \in \text{R}^{H \times W \times 3}$ Where $H$ = 64 and $W$ = 256 , we first use a 45-layer ResNet and a two-layer bidirectional LSTM~\cite{DBLP:journals/pami/GravesLFBBS09} to encode $I$ into a feature map $h^{(v)} \in \text{R}^{1 \times T \times D}$\cite{DBLP:journals/pami/ShiYWLYB19} where $T$ = 25 and $D$ = 512. We can squeeze the height dimension and the dimension of $h^{(v)}$ becomes $T \times D$. The CTC decoder\cite{DBLP:conf/icml/GravesFGS06}, which consists of a two-layer bidirectional LSTM and a softmax layer, takes $h^{(v)}$ as input and outputs a sequence of probability distribution $P_{c}$, and then we can get a coarse prediction $\mathcal{T}_c$ from $P_{c}$ by selecting the most probable characters.

\subsection{Visual Decoder}
The visual decoder is the combination of an attention unit and a GRU unit for extracting character-level visual features.

At time step $t \in [0,T]$, we concatenate the character-level semantic feature $s_t^{(s)}$ (detailed in Section \ref{sm}) and a learnable step embedding $e_{t}$ as the query to get the attention weights on the local visual feature sequence:
\begin{equation}
  \label{alpha1}
  \displaystyle{c_{t,i}=W^{T}\text{tanh}\left(U\left[s_t^{(s)}, e_{t}\right]+Vh_{i}^{(v)}+b\right)}
\end{equation}
\begin{equation}
  \label{alpha2}
  \displaystyle{\alpha_{t,i}=\text{exp}(c_{t,i})/\sum\limits_{i^{'}=1}^T \text{exp}(c_{t,i^{'}})}
\end{equation}

When generating attentional weights $\alpha_{t}$, considering $s_t^{(s)}$ contains semantic information built in the semantic module and this information can make visual decoder attend to feature map's corresponding part more accurately (analyzed in Section \ref{abls}), we choose to use the semantic feature $s_t^{(s)}$ instead of the visual feature $s_{t-1}^{(v)}$. The latter is the common usage in previous attention-based methods.

Obviously, $\alpha_{t}$ can be considered as a location mask on $h^{(v)}$ at current step, hence the attention unit carries out the character feature alignment. The visual glimpse and the character feature are:
\begin{equation}
  \label{glimpse}
  \displaystyle{g_{t}=\sum\limits_{i=1}^{T}\alpha_{t,i}s_{i}}
\end{equation}
\begin{equation}
  \label{svt}
  \displaystyle{s_t^{(v)}=\text{GRU}\left(s_{t-1}^{(v)},g_t\right)}
\end{equation}

It is worth noticing that we do not use the last time step's predicted character $y_{t-1}$ here because bringing in $y_{t-1}$ will lead the visual decoder to capture the dependencies between output characters and build extra semantic information, which is supposed to be built only in the semantic module.

\subsection{Semantic Module}\label{sm}
Our semantic module is designed for word-level language modeling.

\textbf{Semantic Encoder} (SE) first maps a text sequence with length T to $s_d \in \text{R}^{T \times 256}$ with a character embedding layer, and then $s_d$ is fed into a two-layer bidirectional GRU, a linear function and an averaging operation (average across the T dimension) to get the global semantic embedding $s_g \in \text{R}^{256}$.

\textbf{Semantic Decoder} (SD) is designed to generate semantic hidden states step by step. The global semantic embedding $s_g$ is mapped to an initial semantic hidden state $s_0^{(s)}$ and a common word embedding $e_w$ by two linear layers respectively. At time step $t$, the new hidden state $s_t^{(s)}$ is calculated by using the predicted character in the previous step $y_{t-1}$, the previous hidden state $s_{t-1}^{(s)}$ and the word embedding $e_w$ as following:
\begin{equation}
\label{sem_gru}
\displaystyle{s_t^{(s)}=\text{GRU}\left(s_{t-1}^{(s)}, \left[f(y_{t-1}),e_w\right]\right)}
\end{equation}

\subsubsection{Pre-training: a Word Correction Task}\label{pretrain}
To make the semantic module learn word-level semantics beyond the limited-size vocabulary in the training image set, we take a language modeling task which is to simply correct words that might have spelling errors. SE takes a string as input and SD output the corresponding correct word. Our training vocabulary is mainly derived from Synth90K~\cite{DBLP:journals/ijcv/JaderbergSVZ16} which contains 90k words. Besides, we add some random digit numbers to enrich the vocabulary.

When preparing the input string, we follow some specially-designed rules to simulate the coarse text prediction by CTC decoder as real as possible. Specifically, we interrupt a word by exploiting 3 probabilistic operations on a random character: replacement (40\%), insertion (10\%) and deletion (15\%). We replace a character to another one based on the visual similarity matrix $\mathcal{S}\in \mathbb{R}^{C\times C}$, which is calculated by: $\mathcal{S}_0=\text{cos}(W_{ctc},W_{ctc})$, $\mathcal{S}_0[i,i]=0$, $\mathcal{S}=\text{softmax}(k*\mathcal{S}_0)$ successively,
where $W_{ctc}$ is the weights of the classifier in the CTC decoder which is an approximate visual metric for characters, $\text{Cos}$ is the cosine similarity function, $C$ is the number of classes, and $k$ is a hyper-parameter set to 3 empirically. 

\begin{table*}[t]
   \caption{Accuracies of models trained on datasets with small size of vocabulary and samples.}
   \centering
   \begin{tabular}{|l|c|c|c|c|c|c|c|}
   \hline
   \multirow{2}*{Method}& \multirow{2}*{Training Data}& \multicolumn{3}{c|}{IIIT5K}&  \multicolumn{3}{c|}{IC15}\\
   \cline{3-8}
    & & InVoc & OutVoc & Total & InVoc & OutVoc & Total\\
   \hline
   \hline
   No. of images & \multirow{5}*{Synth9K}& 263 & 2737 & 3000 & 136 & 1675 & 1811\\
   \cline{0-0}\cline{3-8}
   Aster & & \textbf{90.1} & 30.0 & 35.3 & \textbf{76.5} & 19.0 & 23.3\\
   Aster$^\star$ & & 88.2 & 67.6 & 69.4 & 69.1 & 57.6 & 58.4\\
   VSDN & & 76.8 & 57.0 & 58.8 & 66.2 & 47.7 & 49.1\\
   VSDN$^\dagger$ & & 87.8 & \textbf{71.3} & \textbf{72.7} & 75.7 & \textbf{65.9} & \textbf{66.6}\\
   \hline
   No. of images & \multirow{5}*{Synth18K}& 468 & 2532 & 3000 & 251 & 1560 & 1811\\
   \cline{0-0}\cline{3-8}
   Aster & & \textbf{88.7} & 54.4 & 59.7 & \textbf{80.9} & 46.3 & 51.1\\
   Aster$^\star$ & & 83.1 & 70.3 & 72.3 & 63.7 & 59.0 & 59.7\\
   VSDN & & 85.7 & 67.7 & 70.5 & 72.1 & 56.9 & 59.0\\
   VSDN$^\dagger$ & & 84.6 & \textbf{75.2} & \textbf{76.7} & 77.3 & \textbf{67.6} & \textbf{68.9}\\
   \hline
   No. of images & \multirow{5}*{Synth45K}& 1231 & 1769 & 3000 & 742 & 1069 & 1811\\
   \cline{0-0}\cline{3-8}
   Aster & & 86.2 & 71.5 & 77.5 & 78.2 & 59.8 & 67.3\\
   Aster$^\star$ & & 82.8 & 74.6 & 78.0 & 67.4 & 59.4 & 62.7\\
   VSDN & & \textbf{87.7} & 75.7 & 80.7 & \textbf{80.7} & 63.5 & 70.6\\
   VSDN$^\dagger$ & & 86.3 & \textbf{80.2} & \textbf{82.7} & 80.1 & \textbf{68.1} & \textbf{73.0}\\
   \hline
   No. of images & \multirow{5}*{Synth90K}& 2415 & 585 & 3000 & 1420 & 391 & 1811\\
   \cline{0-0}\cline{3-8}
   Aster & & 87.0 & 60.5 & 81.9 & 78.0 & 43.7 & 70.6\\
   Aster$^\star$ & & 83.6 & 65.5 & 80.1 & 73.0 & 49.9 & 68.0\\
   VSDN & & \textbf{87.8} & 62.4 & 82.9 & 77.6 & 48.7 & 71.2\\
   VSDN$^\dagger$ & & 87.6 & \textbf{70.1} & \textbf{84.2} & \textbf{80.1} & \textbf{53.7} & \textbf{74.4}\\
   \hline
   \end{tabular}
   \label{results_oov}
 \end{table*}

 \begin{table}[h]
   \caption{Accuracies of models trained on IIIT5K-2000}
   \centering
   \begin{tabular}{|l|ccc|}
   \hline
   \multirow{3}*{Method}& \multicolumn{3}{c|}{Training Data: IIIT5K-2000}\\
   & \multicolumn{3}{c|}{Testing Data: IIIT5K-3000}\\
   \cline{2-4}
   & InVoc(1435) & OutVoc(1565) & Total(3000)\\
   \hline\hline
   Aster & 68.6 & 0.5 & 33.1\\
   Aster$^\star$ & 78.3 & 34.0 & 55.2\\
   VSDN & \textbf{86.5} & 18.6 & 51.1\\
   VSDN$^\dagger$ & 85.5 & \textbf{36.9} & \textbf{60.1}\\
   \hline
   \end{tabular}
   \label{iiit2k}
 \end{table}

\subsection{Fusion Block}
Both the visual information contained in the visual feature $s_{t}^{(v)}$ and the semantic information contained in the semantic feature $s_{t}^{(s)}$ are important for our model to make a precise prediction, so we combine $s_{t}^{(v)}$ with $s_{t}^{(s)}$ by concatenating them and then use a linear function to get the current-step symbol $y_{t}$ as following:
\begin{equation}
\label{eq9}
\displaystyle{p(y_{t})=\text{softmax}(W[s_{t}^{(v)}, s_{t}^{(s)}]+b)}
\end{equation}
\begin{equation}
\label{eq10}
\displaystyle{y_{t} \sim p(y_{t})}
\end{equation}

\subsection{Loss Functions}\label{lf}
The overall loss function consists of four parts, which is defined as following:
\begin{equation}
\label{eq11}
\displaystyle{\mathcal{L}=\lambda _{ctc} \mathcal{L}_{ctc} + \lambda _{v} \mathcal{L}_{v} + \lambda _{s} \mathcal{L}_{s} + \lambda _{f} \mathcal{L}_{f}}
\end{equation}
where $\mathcal{L}_{ctc}$ is the CTC loss function, $\mathcal{L}_{v}$, $\mathcal{L}_{s}$, $\mathcal{L}_{f}$ are the cross-entropy loss function performed on the visual decoder's output, the semantic decoder's output, and the final output. The labels are all the same.
$\lambda_{ctc}$, $\lambda_{v}$, $\lambda_{s}$, $\lambda_{f}$ are hyper-parameters to control the trade-off of the four terms. In our experiments, $\lambda_{c}$, $\lambda_{v}$, $\lambda_{f}$ are set to 1.0, and $\lambda_{s}$ is set to 0.2.

\section{Experiments}
In this section, we firstly conduct experiments to validate the effectiveness of our proposed method in alleviating the problem of vocabulary reliance. 
Next, we compare our method with previous state-of-the art methods on several public benchmark datasets.
Lastly, we make ablation studies to show the effectiveness of several components.

\subsection{Datasets}
\textbf{Synth90K}~\cite{DBLP:journals/ijcv/JaderbergSVZ16} and \textbf{SynthText}~\cite{Synthtext} are the popular training dataset, containing 9 million and 8 million synthetic text line images respectively. 
\textbf{IIIT5K-Words (IIIT5K)}~\cite{IIIT5K} consists of 2000 training images and 3000 testing images. Besides, 
\textbf{Street View Text (SVT)}~\cite{SVT} (647), \textbf{ICDAR2013 (IC13)}~\cite{IC13} (1015), \textbf{ICDAR2015 (IC15)}~\cite{IC15} (1811),  \textbf{SVT-Perspective (SVTP)}~\cite{SVTP} (645) and \textbf{CUTE80 (CUTE)}~\cite{CUTE} (288) are common benchmark datasets for model evaluation.

\subsection{Implementation Details}
The number of classes to be recognized is 39, including 26 lower-case letters, 10 digits and 3 special symbols: end of sequence (EoS), unknown (UKN) and padding (PAD).
For fair comparison, we use the 2 synthetic datasets and their augmented version released by SRN\cite{DBLP:conf/cvpr/YuLZLHLD20} as our training data. We choose Adadelta~\cite{zeiler2012adadelta} as the optimizer to train for 6 epochs with learning rate 1.0 which decays to 0.1 and 0.01 at the 4th and 5th epoch respectively. The batch size is set to 1024 and all experiments are implemented on two GeForce-GTX-1080-Ti graphics cards.

\subsection{Alleviating Vocabulary Reliance}
To validate the effectiveness of VSDN trained on datasets with small-sized vocabulary and image samples, we construct several sub-datasets of Synth90K as training data, i.e., Synth9K, Synth18K, Synth45K by randomly choosing 10\%, 20\%, 50\% vocabularies of Synth90K. Same as Synth90K, every word has 100 image samples.

We adopt the strong popular model Aster~\cite{DBLP:journals/pami/ShiYWLYB19} as the baseline, and Aster without the language model (Aster$^\star$, simply remove the $y_{t-1}$ in its decoder) to show the double-sword effect directly. VSDN$^\dagger$ uses pre-trained parameters of the semantic module to further demonstrate the effectiveness of the pre-training task.

\begin{figure}[h]
   \begin{center}
   \includegraphics[scale=0.45]{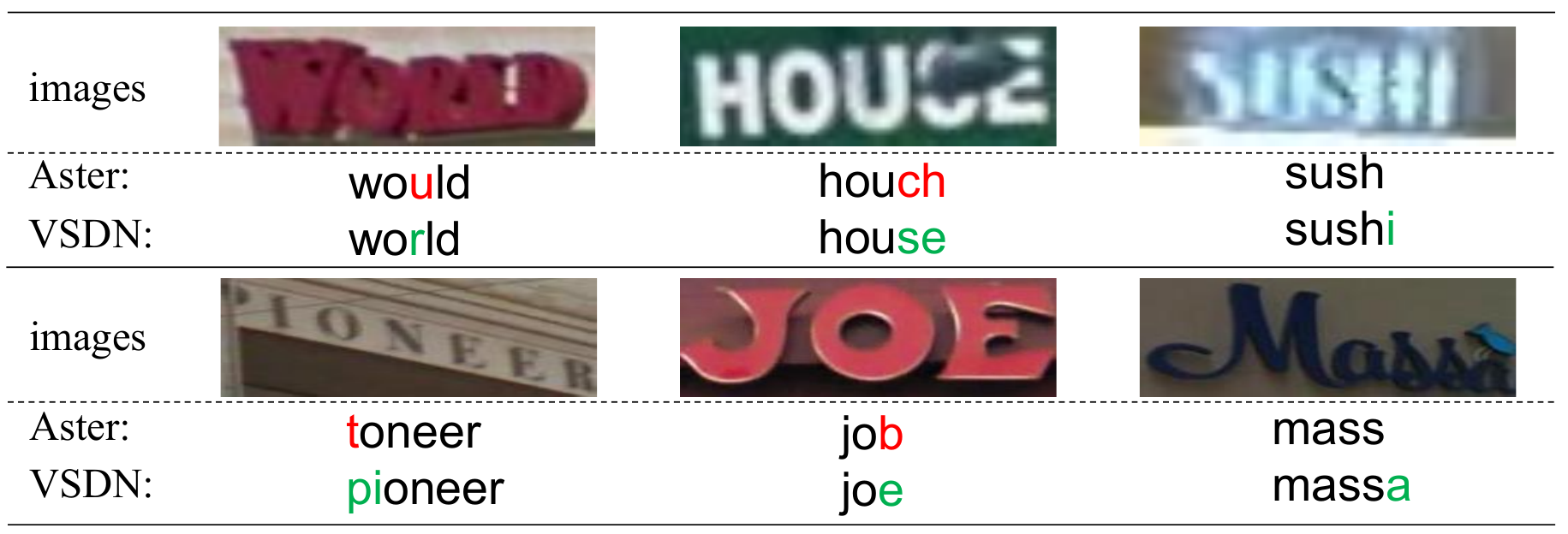}
   \end{center}
   \caption{Some examples for comparison between VSDN and Aster}
   \label{right_case}
\end{figure}

\begin{figure}[h]
   \begin{center}
   \includegraphics[scale=0.4]{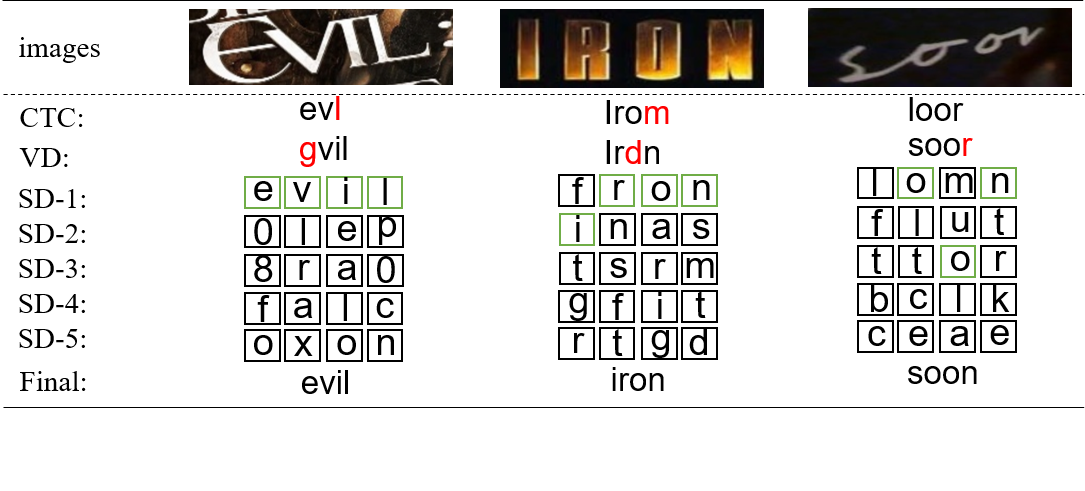}
   \end{center}
   \caption{Examples to show the predictions from different components}
   \label{components}
\end{figure}

For each group of experiment, we first make sure the training vocabulary, and then split the test set into 2 parts: samples with texts in vocabulary and out of vocabulary. The respective and total accuracy are given to measure the performance.

The results are shown in \tab{results_oov}.
After removing the implicit language model, Aster$^\star$ performs better than Aster when training vocabulary is insufficient, which means the coupling modeled semantic information is harmful since it will increase the bias towards the words in training vocabulary.

Our VSDN without pre-training is similar to Aster, but performs better, which indicates that the decoupling structure is less affected by the small-sized training vocabulary. Moreover, VSDN with parts of parameters pre-trained (VSDN$^\dagger$) gets the best accuracy. The simple language modeling task largely enhances our model's capability of constructing semantic information and makes our model more robust against the insufficiency of training vocabulary. 

We also make experiments on a real dataset IIIT5K using only its 2000 real training samples. As shown in \tab{iiit2k}, the conclusion is the same.

\begin{table*}
   \caption{Comparisons on public benchmarks}
   \centering
   \begin{tabular}{|c|c|c|c|c|c|c|}
   \hline
   Method& IIIT5K& SVT& IC13& IC15& SVTP& CUTE\\
   \hline\hline
   CNN\cite{DBLP:journals/ijcv/JaderbergSVZ16} & - & 80.7& 90.8 & - & - & -\\
   CRNN\cite{DBLP:journals/pami/ShiBY17} & 81.2 & 82.7& 89.6 & - & - & -\\
   RRN\cite{DBLP:conf/cvpr/LeeO16} & 78.4 & 80.7 & 90.0 & - & - & -\\
   FAN\cite{DBLP:conf/iccv/ChengBXZPZ17} & 87.4 & 85.9 & 93.3 & 70.6 & - & -\\
   AON\cite{DBLP:conf/cvpr/ChengXBNPZ18} & 87.0 & 82.8& - & 68.2 & 73.0 & 76.8\\
   ACE\cite{DBLP:conf/cvpr/XieHZJLX19} & 82.3 & 82.6 & 89.7 & 68.9 & 70.1 & 82.6\\
   FCN\cite{DBLP:conf/aaai/LiaoZWXLLYB19} & 91.9 & 86.4 & 91.3 & - & - & -\\
   ScRN\cite{DBLP:conf/iccv/YangGLHBBYB19} & 94.4 & 88.9 & 93.9 & 78.7 & 80.8 & 87.5\\
   SAR\cite{DBLP:conf/aaai/00130SZ19} & 91.5 & 84.5 & 91.0 & 69.2 & 76.4 & 83.3\\
   ESIR\cite{DBLP:conf/cvpr/ZhanL19} & 93.3 & 90.2 & 91.3 & 76.9 & 79.6 & 83.3\\
   TextScanner\cite{DBLP:conf/aaai/WanHCBY20} & 93.9 & 90.1 & 92.9 & 79.4 & 84.3 & 83.3\\
   DAN\cite{wang2020decoupled} & 94.3 & 89.2 & 93.9 & 74.5 & 80.0 & 84.4\\
   LAL\cite{DBLP:conf/mm/ZhengQWB20} & \textbf{95.0} & 89.8 & 95.1 & 79.0 & 82.9 & 87.8\\
   SEED\cite{DBLP:conf/cvpr/QiaoZYZ020} & 93.8 & 89.6 & 92.8 & 80.0 & 81.4 & 83.6\\
   SRN\cite{DBLP:conf/cvpr/YuLZLHLD20} & 94.8 & 91.5 & \textbf{95.5} & 82.7 & 85.1 & \textbf{87.8}\\
   \hline
   Aster (Baseline) \cite{DBLP:journals/pami/ShiYWLYB19} & 93.4 & 89.5 & 91.8 & 76.1 & 78.5 & 79.5\\
   VSDN (Ours) & 94.4 & \textbf{92.3} & 93.5 & \textbf{84.5} & \textbf{85.3} & 85.1\\
   \hline
   \end{tabular}
   \label{results_sota}
 \end{table*}

\subsection{Comparison with State-of-the-art}
We compare our VSDN with previous state-of-the-art methods in \tab{results_sota}. Our proposed method VSDN achieves the best results on SVT, SVTP and IC15 and competitive results on the rest of the datasets. Note that LAL\cite{DBLP:conf/mm/ZhengQWB20} uses more curved synthetic text images to train the model, thus it is not fair to compare directly. 

\begin{table}
   \caption{Performances of different components}
   \centering
   \begin{tabular}{|c|c|c|c|c|c|c|}
   \hline
   Method& IIIT5K& SVT& IC13& IC15& SVTP\\
   \hline\hline
   VSDN (CTC) & 92.1 & 89.0 & 90.9 & 80.6 & 78.3\\
   \hline
   VSDN (VD) & 93.9 & 91.8 & 92.9 & 83.8 & 84.2 \\
   \hline
   VSDN (SD$_{top1}$) & 84.7 & 90.4 & 91.2 & 75.8 & 81.6\\
   VSDN (SD$_{top3}$) & 90.3 & 92.4 & 93.7 & 82.4 & 85.4\\
   VSDN (SD$_{top5}$) & 92.4 & \textbf{94.0} & \textbf{95.2} & \textbf{85.8} & \textbf{88.1}\\
   \hline
   VSDN (Final) & \textbf{94.4} & 92.3 & 93.5 & 84.5 & 85.3\\
   \hline
   \end{tabular}
   \label{results_cvsp}
 \end{table}

For regular datasets, compared with our baseline Aster~\cite{DBLP:journals/pami/ShiYWLYB19}, VSDN improves 1.0\% on IIIT5K (from 93.4\% to 94.4\%), 2.8\% on SVT (from 89.5\% to 92.3\%) and 1.7\% on IC13 (from 91.8\% to 93.5\%). 

VSDN also performs well on irregular datasets. VSDN improves 8.4\% on IC15 (from 76.1\% to 84.5\%), 6.8\% on SVTP (from 78.5\% to 85.3\%) and 5.6\% on CUTE (from 79.5\% to 85.1\%) when compared with Aster~\cite{DBLP:journals/pami/ShiYWLYB19}. 

Although many images in SVT are severely corrupted by noise, blur, and low resolution and many images in IC15 suffer in heavy perspective distortions, our method still largely outperforms previous methods on the two datasets. We believe it is owing to that our VSDN can construct accurate semantic information, which is robust for visual defects. Examples are shown in \fig{right_case}.

\subsection{Component Analysis}
To be clear how the visual and semantic features contribute to the final recognition, we also evaluate the performance of each component that is supervised during training. Specially, for the prediction of semantic decoder, we adopt top-$k$ accuracy since the word correction task usually has more than 1 answer (e.g., big VS bug VS bag).

As shown in \tab{results_cvsp}, CTC decoder has relatively low accuracy. But it is enough for semantic module because the character error rate is adequately low. With the function of word correction, the semantic decoder has surpassed the CTC decoder by its top-$k$ accuracy, which proves that our VSDN has the ability of semantic modeling. The final performance is better than the pure visual decoder, showing that the feature fusion is necessary. Note that the high accuracy of the visual decoder is not just owing to the visual learning, but the joint semantic learning and character feature alignment (See Sec. \ref{cfa}). \fig{components} shows some cases to understand the above analysis intuitively. The results of semantic decoder are presented by the top-5 prediction.

\subsection{Ablation Study}\label{abls}

\subsubsection{Effectiveness of Visual Loss and Semantic Loss}
In the training stage, $\mathcal{L}_{v}$ and $\mathcal{L}_{s}$ are responsible for the supervision of the output of visual decoder and semantic decoder respectively (detailed in Section \ref{lf}). We conduct experiments to evaluate the effectiveness of them. As shown in \tab{vsloss}, the two loss terms matter much and removing any of them will lead to a drop in accuracy, which indicates that it is essential for both the visual and semantic parts to be supervised with labels.

\begin{figure}[h]
   \begin{center}
   \includegraphics[scale=0.5]{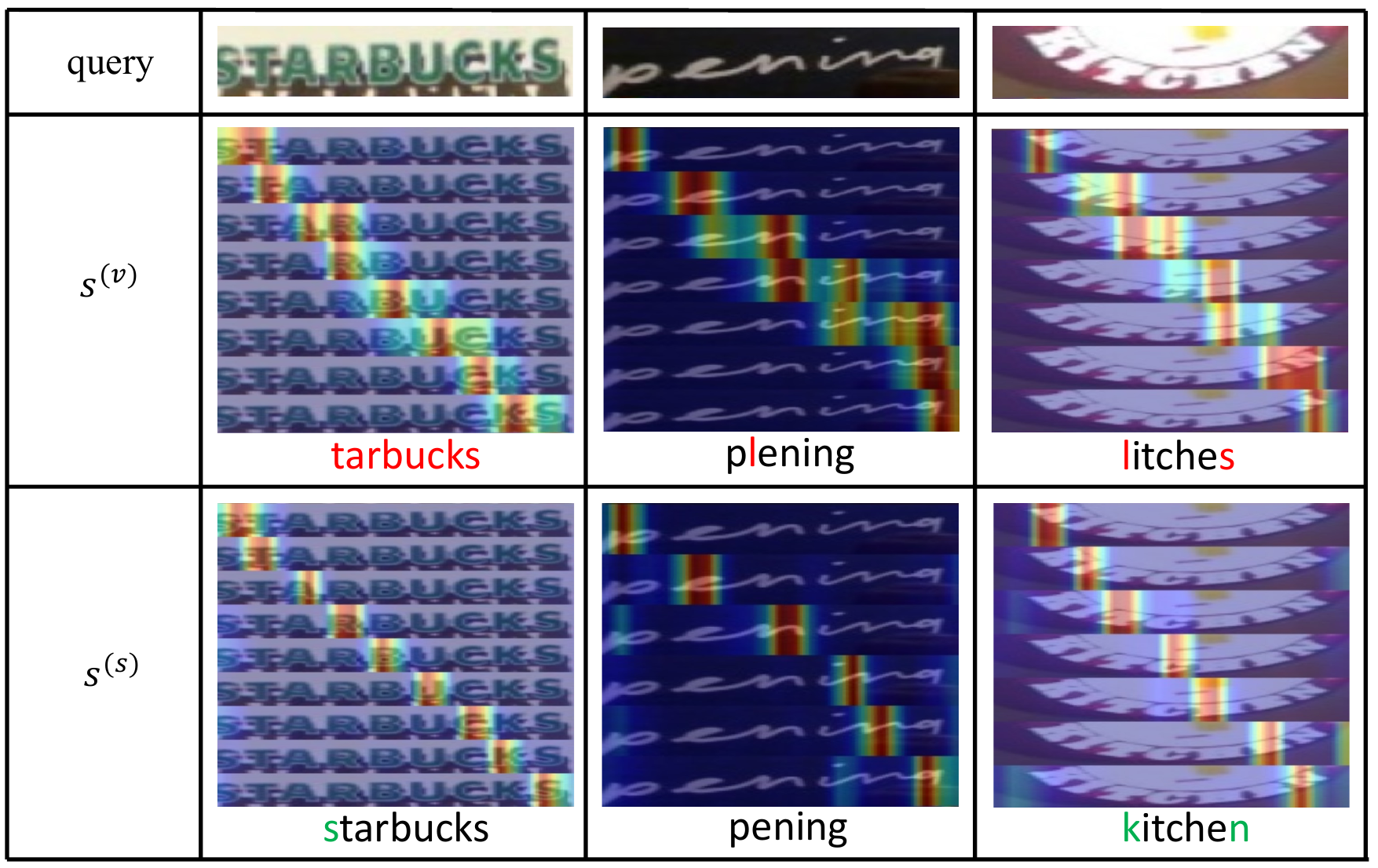}
   \end{center}
   \caption{Comparison of attention map by different queries.}
   \label{hidden}
\end{figure}

\begin{table}[t]
    \centering
      \makeatletter\def\@captype{table}\makeatother\caption{The effect of visual and semantic loss for the training of VSDN.}
      \label{vsloss}
      \begin{tabular}{|m{.18\columnwidth}<{\centering}|m{.05\columnwidth}<{\centering}m{.05\columnwidth}<{\centering}|m{.10\columnwidth}<{\centering}|m{.11\columnwidth}<{\centering}|m{.13\columnwidth}<{\centering}|}
         \hline
         Method & $\mathcal{L}_{v}$ & $\mathcal{L}_{s}$ & SVT & IC13 & SVTP\\
         \hline\hline
         \multirow{4}*{VSDN} &  &  & 90.6 & 92.7 & 82.9\\
          & \checkmark &  & 91.5 & 92.3 & 83.1\\
          &  & \checkmark & 91.5 & 93.0 & 83.3\\
          & \checkmark & \checkmark & \textbf{92.3} & \textbf{93.5} & \textbf{85.3}\\
         \hline
      \end{tabular}
\end{table}

\begin{table}[t]
     \centering
          \makeatletter\def\@captype{table}\makeatother\caption{The comparison of different queries in visual decoder.}
          \label{query}
          \begin{tabular}{|m{.16\columnwidth}<{\centering}|m{.12\columnwidth}<{\centering}|m{.12\columnwidth}<{\centering}|m{.15\columnwidth}<{\centering}|}
            \hline
            query & SVT & IC15 & CUTE\\
            \hline\hline
            $s^{(v)}_{t-1}$ & 90.1 & 81.8 & 83.2\\
            $s^{(s)}_t$ & \textbf{92.3} & \textbf{84.5} & \textbf{85.1}\\
            \hline
        \end{tabular}
\end{table}

\subsubsection{Query in Character Feature Alignment}\label{cfa}
In the phase of attention-based character feature alignment, the semantic feature of the current step $s^{(s)}_t$ calculated from the semantic decoder is utilized as the query. In Aster~\cite{DBLP:journals/pami/ShiYWLYB19}, they used the previous hidden state $s^{(v)}_{t-1}$ as query.

To explore the different effects between these two hidden states, we conduct experiments by choosing each hidden state as the query and the quantitative results are shown in \tab{query}. Compared with visual hidden state $s^{(v)}_{t-1}$, taking semantic hidden state $s^{(s)}_t$ as the query achieves better performance. We argue that $s^{(v)}_{t-1}$ is prone to be biased by the limited and noisy training vocabulary since the semantic feature is implicitly modeled in a coupling way. Accurate alignment needs accurate semantic feature. \fig{hidden} shows some visualization examples.

\section{Conclusion}
In this paper, we propose a novel Visual-Semantic Decoupling Network to learn the character-level visual and semantic features independently. Taking the advantage of decoupling, we can pre-train the semantic module by a word correction task with inexpensive text data. Our method achieves state-of-the-art or competitive results on several public benchmarks, and shows great superiority against the baseline when the taining data only has a small size of samples and vocabulary, which alleviates the problem of vocabulary reliance.

{\small
\bibliographystyle{ieee_fullname}
\bibliography{egbib}

\begin{thebibliography}{10}\itemsep=-1pt

\bibitem{DBLP:journals/corr/BahdanauCB14}
Dzmitry Bahdanau, Kyunghyun Cho, and Yoshua Bengio.
\newblock Neural machine translation by jointly learning to align and
  translate.
\newblock In {\em ICLR}, 2015.

\bibitem{DBLP:journals/ijon/ChenWZJL20}
Xiaoxue Chen, Tianwei Wang, Yuanzhi Zhu, Lianwen Jin, and Canjie Luo.
\newblock Adaptive embedding gate for attention-based scene text recognition.
\newblock {\em Neurocomputing}, 381:261--271, 2020.

\bibitem{DBLP:conf/iccv/ChengBXZPZ17}
Zhanzhan Cheng, Fan Bai, Yunlu Xu, Gang Zheng, Shiliang Pu, and Shuigeng Zhou.
\newblock Focusing attention: Towards accurate text recognition in natural
  images.
\newblock In {\em ICCV}, pages 5086--5094, 2017.

\bibitem{DBLP:conf/cvpr/ChengXBNPZ18}
Zhanzhan Cheng, Yangliu Xu, Fan Bai, Yi Niu, Shiliang Pu, and Shuigeng Zhou.
\newblock {AON:} towards arbitrarily-oriented text recognition.
\newblock In {\em CVPR}, pages 5571--5579, 2018.

\bibitem{DBLP:conf/icml/GravesFGS06}
Alex Graves, Santiago Fern{\'{a}}ndez, Faustino~J. Gomez, and J{\"{u}}rgen
  Schmidhuber.
\newblock Connectionist temporal classification: labelling unsegmented sequence
  data with recurrent neural networks.
\newblock In {\em ICML}, volume 148, pages 369--376, 2006.

\bibitem{DBLP:journals/pami/GravesLFBBS09}
Alex Graves, Marcus Liwicki, Santiago Fern{\'{a}}ndez, Roman Bertolami, Horst
  Bunke, and J{\"{u}}rgen Schmidhuber.
\newblock A novel connectionist system for unconstrained handwriting
  recognition.
\newblock {\em TPAMI}, 31(5):855--868, 2009.

\bibitem{Synthtext}
Ankush Gupta, Andrea Vedaldi, and Andrew Zisserman.
\newblock Synthetic data for text localisation in natural images.
\newblock In {\em CVPR}, pages 2315--2324, 2016.

\bibitem{DBLP:conf/interspeech/HoriWZC17}
Takaaki Hori, Shinji Watanabe, Yu Zhang, and William Chan.
\newblock Advances in joint ctc-attention based end-to-end speech recognition
  with a deep {CNN} encoder and {RNN-LM}.
\newblock In {\em INTERSPEECH}, pages 949--953, 2017.

\bibitem{DBLP:conf/aaai/HuCHYL20}
Wenyang Hu, Xiaocong Cai, Jun Hou, Shuai Yi, and Zhiping Lin.
\newblock {GTC:} guided training of {CTC} towards efficient and accurate scene
  text recognition.
\newblock In {\em AAAI}, pages 11005--11012, 2020.

\bibitem{DBLP:journals/ijcv/JaderbergSVZ16}
Max Jaderberg, Karen Simonyan, Andrea Vedaldi, and Andrew Zisserman.
\newblock Reading text in the wild with convolutional neural networks.
\newblock {\em IJCV}, 116(1):1--20, 2016.

\bibitem{IC15}
Dimosthenis Karatzas, Lluis Gomez{-}Bigorda, Anguelos Nicolaou, Suman~K. Ghosh,
  Andrew~D. Bagdanov, Masakazu Iwamura, Jiri Matas, Lukas Neumann,
  Vijay~Ramaseshan Chandrasekhar, Shijian Lu, Faisal Shafait, Seiichi Uchida,
  and Ernest Valveny.
\newblock {ICDAR} 2015 competition on robust reading.
\newblock In {\em ICDAR}, pages 1156--1160, 2015.

\bibitem{IC13}
Dimosthenis Karatzas, Faisal Shafait, Seiichi Uchida, Masakazu Iwamura,
  Lluis~Gomez i Bigorda, Sergi~Robles Mestre, Joan Mas, David~Fern{\'{a}}ndez
  Mota, Jon Almaz{\'{a}}n, and Llu{\'{\i}}s{-}Pere de~las Heras.
\newblock {ICDAR} 2013 robust reading competition.
\newblock In {\em ICDAR}, pages 1484--1493, 2013.

\bibitem{DBLP:conf/cvpr/LeeO16}
Chen{-}Yu Lee and Simon Osindero.
\newblock Recursive recurrent nets with attention modeling for {OCR} in the
  wild.
\newblock In {\em CVPR}, pages 2231--2239, 2016.

\bibitem{DBLP:conf/aaai/00130SZ19}
Hui Li, Peng Wang, Chunhua Shen, and Guyu Zhang.
\newblock Show, attend and read: {A} simple and strong baseline for irregular
  text recognition.
\newblock In {\em AAAI}, pages 8610--8617, 2019.

\bibitem{DBLP:conf/aaai/LiaoZWXLLYB19}
Minghui Liao, Jian Zhang, Zhaoyi Wan, Fengming Xie, Jiajun Liang, Pengyuan Lyu,
  Cong Yao, and Xiang Bai.
\newblock Scene text recognition from two-dimensional perspective.
\newblock In {\em AAAI}, pages 8714--8721, 2019.

\bibitem{IIIT5K}
Anand Mishra, Karteek Alahari, and C.~V. Jawahar.
\newblock Scene text recognition using higher order language priors.
\newblock In {\em BMVC}, pages 1--11, 2012.

\bibitem{SVTP}
Trung~Quy Phan, Palaiahnakote Shivakumara, Shangxuan Tian, and Chew~Lim Tan.
\newblock Recognizing text with perspective distortion in natural scenes.
\newblock In {\em ICCV}, pages 569--576, 2013.

\bibitem{DBLP:conf/cvpr/QiaoZYZ020}
Zhi Qiao, Yu Zhou, Dongbao Yang, Yucan Zhou, and Weiping Wang.
\newblock {SEED:} semantics enhanced encoder-decoder framework for scene text
  recognition.
\newblock In {\em CVPR}, pages 13525--13534, 2020.

\bibitem{CUTE}
Anhar Risnumawan, Palaiahnakote Shivakumara, Chee~Seng Chan, and Chew~Lim Tan.
\newblock A robust arbitrary text detection system for natural scene images.
\newblock {\em Expert Systems with Applications}, 41(18):8027--8048, 2014.

\bibitem{DBLP:journals/pami/ShiBY17}
Baoguang Shi, Xiang Bai, and Cong Yao.
\newblock An end-to-end trainable neural network for image-based sequence
  recognition and its application to scene text recognition.
\newblock {\em TPAMI}, 39(11):2298--2304, 2017.

\bibitem{DBLP:journals/pami/ShiYWLYB19}
Baoguang Shi, Mingkun Yang, Xinggang Wang, Pengyuan Lyu, Cong Yao, and Xiang
  Bai.
\newblock {ASTER:} an attentional scene text recognizer with flexible
  rectification.
\newblock {\em TPAMI}, 41(9):2035--2048, 2019.

\bibitem{DBLP:conf/nips/VaswaniSPUJGKP17}
Ashish Vaswani, Noam Shazeer, Niki Parmar, Jakob Uszkoreit, Llion Jones,
  Aidan~N. Gomez, Lukasz Kaiser, and Illia Polosukhin.
\newblock Attention is all you need.
\newblock In {\em NIPS}, pages 5998--6008, 2017.

\bibitem{DBLP:conf/aaai/WanHCBY20}
Zhaoyi Wan, Minghang He, Haoran Chen, Xiang Bai, and Cong Yao.
\newblock Textscanner: Reading characters in order for robust scene text
  recognition.
\newblock In {\em AAAI}, pages 12120--12127, 2020.

\bibitem{DBLP:conf/cvpr/WanZZLY20}
Zhaoyi Wan, Jielei Zhang, Liang Zhang, Jiebo Luo, and Cong Yao.
\newblock On vocabulary reliance in scene text recognition.
\newblock In {\em CVPR}, pages 11422--11431, 2020.

\bibitem{SVT}
Kai Wang, Boris Babenko, and Serge~J. Belongie.
\newblock End-to-end scene text recognition.
\newblock In {\em ICCV}, pages 1457--1464, 2011.

\bibitem{wang2020decoupled}
Tianwei Wang, Yuanzhi Zhu, Lianwen Jin, Canjie Luo, Xiaoxue Chen, Yaqiang Wu,
  Qianying Wang, and Mingxiang Cai.
\newblock Decoupled attention network for text recognition.
\newblock In {\em the AAAI Conference on Artificial Intelligence}, volume~34,
  pages 12216--12224, 2020.

\bibitem{DBLP:conf/cvpr/XieHZJLX19}
Zecheng Xie, Yaoxiong Huang, Yuanzhi Zhu, Lianwen Jin, Yuliang Liu, and Lele
  Xie.
\newblock Aggregation cross-entropy for sequence recognition.
\newblock In {\em CVPR}, pages 6538--6547, 2019.

\bibitem{DBLP:conf/iccv/YangGLHBBYB19}
Mingkun Yang, Yushuo Guan, Minghui Liao, Xin He, Kaigui Bian, Song Bai, Cong
  Yao, and Xiang Bai.
\newblock Symmetry-constrained rectification network for scene text
  recognition.
\newblock In {\em ICCV}, pages 9146--9155, 2019.

\bibitem{DBLP:conf/cvpr/YuLZLHLD20}
Deli Yu, Xuan Li, Chengquan Zhang, Tao Liu, Junyu Han, Jingtuo Liu, and Errui
  Ding.
\newblock Towards accurate scene text recognition with semantic reasoning
  networks.
\newblock In {\em CVPR}, pages 12110--12119, 2020.

\bibitem{zeiler2012adadelta}
Matthew~D Zeiler.
\newblock Adadelta: an adaptive learning rate method.
\newblock {\em arXiv preprint arXiv:1212.5701}, 2012.

\bibitem{DBLP:conf/cvpr/ZhanL19}
Fangneng Zhan and Shijian Lu.
\newblock {ESIR:} end-to-end scene text recognition via iterative image
  rectification.
\newblock In {\em CVPR}, pages 2059--2068, 2019.

\bibitem{DBLP:conf/mm/ZhengQWB20}
Yi Zheng, Wenda Qin, Derry Wijaya, and Margrit Betke.
\newblock {LAL:} linguistically aware learning for scene text recognition.
\newblock In {\em ACM MM}, pages 4051--4059, 2020.

\bibitem{DBLP:journals/access/ZuoSMQJ19}
Ling{-}Qun Zuo, Hong{-}mei Sun, Qi{-}Chao Mao, Rong Qi, and Ruisheng Jia.
\newblock Natural scene text recognition based on encoder-decoder framework.
\newblock {\em {IEEE} Access}, 7:62616--62623, 2019.

\end{thebibliography}
}

\end{document}